\documentclass{article} 
\usepackage{nips14submit_e,times}

\usepackage{amsmath}
\usepackage{amsfonts}
\usepackage{multirow}
\usepackage{url}
\usepackage{graphicx} 

\usepackage{boxedminipage}  
\usepackage{color}
\usepackage{colortbl}
\usepackage{enumitem}

\newcommand{\mention}[1]{\textit{#1}}
\newcommand{\entity}[1]{\texttt{#1}}

\def\bs#1{\boldsymbol{#1}}

\title{Unsupervised Induction of Semantic Roles \\ within a Reconstruction-Error
Minimization Framework}

\author{
Ivan Titov ~~~~~~ Ehsan Khoddam \\
Institute for Logic, Language and Computation \\
Universiteit van Amsterdam\\
1090 GE, Amsterdam, The Netherlands \\
\texttt{ {\{}titov,e.khoddammohammadi{\}}@uva.nl} \\
}

\nipsfinalcopy 

\begin{document}

\maketitle

\begin{abstract}
We introduce a new approach to  unsupervised estimation of feature-rich semantic role labeling models.
Our model consists of two components: (1) an encoding component: a semantic role labeling
model which predicts roles given a rich set of syntactic and lexical features; (2) a reconstruction component:
a tensor factorization model which relies on roles to predict argument fillers. 
When the components are estimated jointly to minimize
errors in argument reconstruction,  the induced roles largely correspond to roles defined in annotated resources.
Our method performs on par with most accurate  role induction methods on English and German, even though, unlike these previous approaches,  we do not
incorporate any prior linguistic knowledge about the languages.
\end{abstract}

\section{Introduction}

Shallow representations of  meaning,  and  semantic  role  labels  in
particular,  have a  long history  in  linguistics \cite{fillmore-68}.
More recently, with an emergence of large annotated resources such as
PropBank \cite{Palmer05} and FrameNet \cite{Baker98}, automatic semantic role
labeling (SRL) has  attracted a lot of
attention~\cite{Gildea02,conll05,conll08,conll09,Das2010}.

Semantic role representations encode the underlying predicate-argument structure of sentences, or,
more specifically, for every predicate in a sentence they identify
a set of arguments and associate each argument with an underlying {\it semantic
role}, such as an agent (an initiator or doer of the action) or a  patient
(an affected entity).
Semantic roles have many potential applications in NLP and have  been shown to
benefit question answering \cite{Shen07,Kaisser07}, textual entailment \cite{Sammons09}, machine translation 
\cite{WuFung2009,LiuGildea2010,Wu11,GaoVogel11MT}, and dialogue systems \cite{basili-ea09,VanDerPlasNaacl09}, among others.

Most current statistical approaches to SRL are supervised, requiring large quantities of human annotated data to estimate 
model parameters.  However, such resources are expensive to create and only available for a small number of languages and domains.  
Moreover, when moved to a new domain, the performance of these models tends to degrade substantially~\cite{Pradhan08}.  The scarcity of 
annotated data has motivated the research into unsupervised learning of semantic representations~\cite{Swier04,Grenager06,Lang10,Lang11a,Lang11b,TitovKlementEacl12,furstenau2012,garg2012}. 
The existing methods have a number of serious shortcomings.  First, they make very strong assumptions, for example, assuming
that arguments  are conditionally independent of each other given the predicate. Second, unlike state-of-the-art supervised parsers, they rely on a very simplistic set of features of a sentence. 
These factors lead to models being insufficiently expressive to capture syntax-semantics interface,  inadequate handling of  language ambiguity and, overall, introduces an upper bound on their performance. 
Moreover,  these approaches are especially problematic for languages with freer word order than English, where richer features are necessary to account for interactions between surface realizations, syntax and semantics. 
For example, the two  most accurate previous models~\cite{TitovKlementEacl12,Lang11a} both
treat the role induction task as clustering of argument signatures: 
an argument signature encode key syntactic properties of an argument realization and consists of a syntactic function of an argument along with additional informations such as argument
position with respect to the predicate.  Though it is possible to design signatures which mostly map to a single role,
this set-up limits oracle performance even for English, and can be quite restrictive for languages with freer word order.
These shortcomings are inherent limitations of the modeling frameworks used in previous work (primarily generative modeling or agglomerative clustering), and cannot be addressed by simply incorporating more features or
relaxing some of the modeling assumptions.


 In this work, we propose a method  for  effective unsupervised estimation of feature-rich models of semantic roles.  We demonstrate that reconstruction-error objectives, which have been shown to be effective primarily for training neural networks, 
are well suited for inducing feature-rich log-linear models of semantics. Our model consists of two components: a log-linear feature rich semantic role labeler and
a tensor-factorization model which captures interaction between semantic roles and argument fillers. When estimated jointly on unlabeled data, roles induced by the model mostly corresponds to roles defined in existing resources by annotators.

Our method rivals the most accurate 
semantic role induction methods on English and German ~\cite{TitovKlementEacl12,Lang11a}. 
Importantly, no prior knowledge about the languages was incorporated in our feature-rich model,
whereas the clustering counterparts relied on language-specific argument signatures. This languages-specific priors were crucial
for the success, for example, using English-specific argument signatures for German with the Bayesian model of~\citet{TitovKlementEacl12} results in a drop of performance from clustering F1 80.9\% to considerably lower 78.3\% (our model yields 81.4\%). 
This confirms the intuition that using richer features helps to capture the syntax-semantics interface in  multi-lingual settings, reducing
the need for using language-specific model engineering, as highly desirable in unsupervised learning. 



The rest of the paper is structured as follows. 
Section 2 begins with a definition of the semantic
role labeling task and discusses some specifics of the unsupervised setting.  In Section 3, we describe our approach, starting with a general motivation and
proceeding to technical details of  the model (Section 3.3) and the learning procedure (Section 3.4). 
Section 4 provides both
evaluation and analysis. 
Finally, additional related work
is presented in Section 5.

\section{Task definition}
\label{sect:task}

The SRL task involves prediction of predicate
argument structure, i.e. both identification of arguments and assignment of labels according to
their underlying semantic role. For example, in the following sentences:

\begin{description}
\item[(a)] $[_{Agent}$ Mary$]$ opened $[_{Patient}$ the door$]$.
\vspace{-1ex}
\item[(b)] $[_{Patient}$ The door$]$ opened.
\vspace{-1ex}
\item[(c)] $[_{Patient}$ The door$]$ was opened $[_{Agent}$ by Mary$]$.
\end{description}
{\it Mary} always takes an agent role 
for the predicate {\it open}, and {\it door}
is always a patient.  

In this work we focus on the labeling stage
of semantic role labeling.
Identification, though an important problem, can be
tackled   with heuristics~\cite{Lang11a,Grenager06,deMarneffe06}, with 
unsupervised techniques~\cite{Abend09} or potentially by using a supervised classifier
trained on a small amount of data.

\section{Approach}

At the core of our approach is a 
statistical model encoding an interdependence between a semantic role structure and its realization in a sentence.
In the unsupervised learning setting,  
sentences, their syntactic representations and argument positions 
(denoted by $x$) are observable whereas the associated semantic roles $\bs{r}$ are latent and need to be induced by the model. 
The idea which underlines much of latent variable modeling is that 
a good  latent representation is the one which helps us to reconstruct $x$. 
In practice, we are not interested in predicting $x$, as $x$ is observable, but rather
interested in inducing appropriate latent representations (i.e. $\bs{r}$). Thus, it is crucial to design the model in such a way that the good $\bs{r}$  (the one  predictive of $x$) indeed encodes roles, rather than some other form of abstraction.

In what follows, we will refer to roles using their names, though, in the unsupervised setting, our method, as any other latent variable model, will not yield human-interpretable labels for them.  
We will use the following sentence as a motivating example in our discussion of the model:
\begin{description}
\item[] $[_{Agent}$ The police$]$ charged $[_{Patient}$ the demonstrators$]$  $[_{Instrument}$ with batons$]$.
\end{description}

The model consists of two components.
The first component is responsible for prediction of argument tuples based on roles and the predicate. 
In our experiments,  in this component, we  represent arguments as lemmas of their lexical heads (e.g., {\it baton} instead of {\it with batons}), and we also
restrict ourselves to only verbal predicates. 
Intuitively, we can think of predicting one argument at a time (see Figure~1(b)): an argument
 (e.g., \mention{demonstrator} in our example) is predicted based on the predicate lemma (\mention{charge}), 
the role assigned to this argument (i.e. \entity{Patient}) and other role-argument pairs ((\entity{Agent}, \mention{police}) and
(\entity{Instrument}, \mention{baton})).  While learning to predict arguments, the inference algorithm will search for role assignments which 
simplify this prediction task as much as possible. Our hypothesis is that these assignments will correspond to 
roles 
 accepted in linguistic theories (or, more importantly, useful in practical applications). 
Why is this hypothesis plausible? 
Primarily because these semantic representations were introduced as an abstraction capturing crucial properties of a relation (or an event). Thus, these representations, rather than surface linguistic details like argument order or syntactic functions, should be crucial for modeling sets of potential argument tuples.
%
%
The reconstruction component is not the only  part of the model.
Crucially, what we referred to above as 
`searching for role assignments to simplify argument prediction'  would actually correspond to learning another component: a 
semantic role labeler which predicts roles relying on a rich set of sentence features. These two components will be estimated jointly in such a way as to minimize errors in recovering arguments.  The role labeler will be the end-product of learning: it will be used to process new sentences, and it will be compared to existing methods in our evaluation.

\subsection{Shortcomings of generative modeling}
The above paragraph can be regarded as our desiderata; now we discuss how to achieve them. 
The  standard way to approach latent variable modeling is to use the generative 
framework:
that is to define a family of joint models $p(x, y | \theta)$ and estimate the parameters $\theta$
by, for example, maximizing likelihood. 
Generative models of semantics~\cite{TitovKlementEacl12,Titov11,connor2013,kawahara2014} necessarily make very strong independence assumptions (e.g., arguments are conditionally independent of each other given the predicate) and use simplistic features of $x$ and $y$. Thus, they cannot meet the desiderata stated above. Importantly, they are also much more simplistic in their assumptions than state-of-the-art supervised role labelers~\cite{Erk2006,johansson2008,Das2010}. 

\begin{figure*}
\label{fig:reconstr}
\begin{center}
\includegraphics[width=\columnwidth]{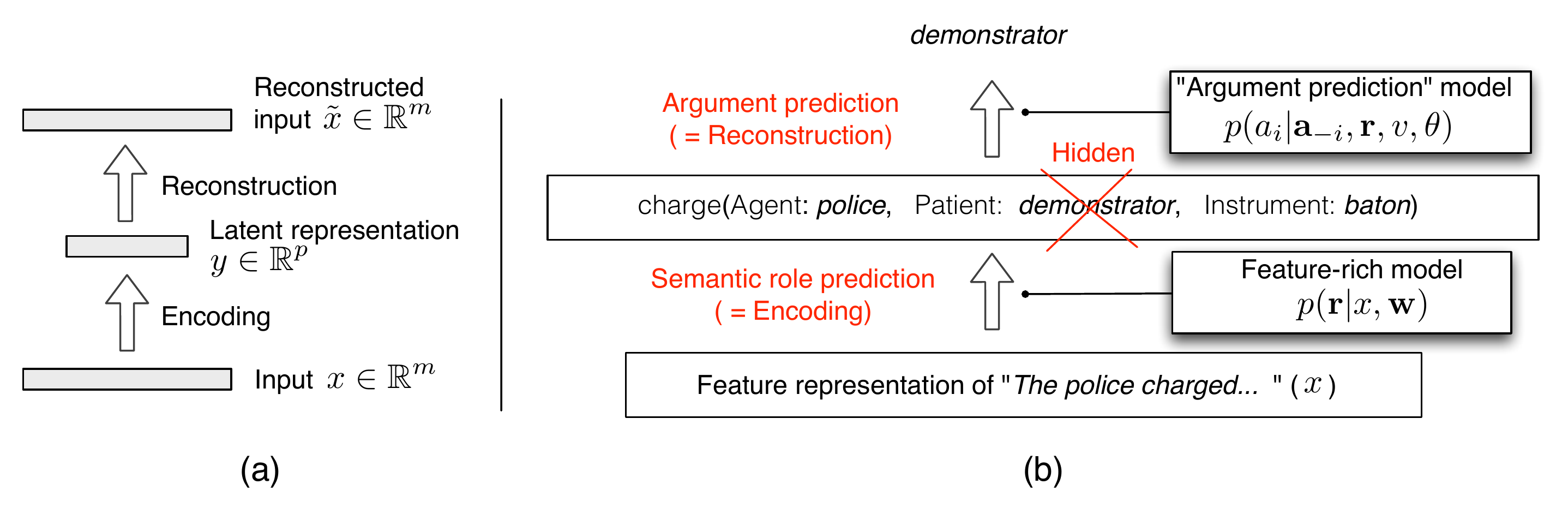}
\caption{(a) An autoencoder from $\mathbb{R}^m$ to $\mathbb{R}^p$ (typically $p < m$).  ~~ (b) Modeling roles within the reconstruction-error minimization framework.
} 
\end{center}
\end{figure*}

\subsection{Reconstruction error minimization}
Generative modeling is not the only way  to learn latent representations.
One alternative, popular in the neural network community, is to instead use  autoencoders and optimize the reconstruction error~\cite{hinton1989,vincent2008}.
In autoencoders, a latent representation $y$ (their hidden layer) is predicted from $x$ by an encoding model  and then this $y$ is used to recover $\tilde{x}$ with a reconstruction model (see Figure~1(a)). Parameters of the encoding and reconstruction components are chosen so as to minimize some form of the reconstruction error, for example, the Euclidean distance
$\Delta(x,\tilde{x}) = ||x-\tilde{x}||_2$.  
Though currently popular only within the deep learning community,
latent variable models other than neural networks can also be trained this way, moreover:
\begin{itemize}
\item the encoding and reconstruction models 
can belong to different model families;
\item the reconstruction component may be focused on recovering a part of $x$ rather than the entire $x$, and, in doing so,  can rely not only on $y$ but on the remaining part of $x$.
\end{itemize}
These observations are crucial as they allow us to implement our desiderata. 
More specifically, the encoding model will be a feature-rich classifier which predicts semantic roles for a sentence, and the reconstruction model is the model which predicts an argument given its role, and given the rest of the arguments and their roles. 
The idea of training linear models with reconstruction error was previously explored by~\citet{daume2009} and very recently by~\citet{ammar2014}. 

\subsection{Modeling semantics within the reconstruction-error framework}
There are several possible ways to translate the ideas above into a specific method, and we consider one  of the simplest instantiations. 
For simplicity, in the discussion (but not in our experiments), we assume that exactly one predicate is realized in each sentence $x$. 
As we mentioned above, we focus on argument
labeling: we assume that arguments $\bs{a}  =
(a_1, \ldots, a_N)$, $a_i \in \mathcal{A}$,
 are known, and only their roles $\bs{r}  = (r_1, \ldots, r_N)$, $r_i \in \mathcal{R}$ need to be induced. 
For the encoder (i.e. the semantic role labeler), we use a log-linear model:
\begin{equation}
\nonumber
p(\bs{r}| x, \bs{w}) \propto  
\exp(\bs{w}^T \bs{g}(x, \bs{r})),
\end{equation}
where $\bs{g}(x, \bs{r})$ is a feature vector 
encoding interactions between sentence $x$ and the semantic
role representation $\bs{r}$. Any model can be used here as long as
the posterior distributions of roles $r_i$  can
be efficiently computed or approximated (we will see why in Section~\ref{sect:learning}).  In our experiments, we used a model which
factorizes over individual arguments (i.e. independent logistic regression classifiers).
  
The reconstruction component predicts an argument (e.g.,  the $i$th argument
$a_i$) given the semantic roles $\bs{r}$,  the predicate  $v$ and other
arguments $\bs{a}_{-i} =  (a_1, \ldots, a_{i-1}, a_{i+1}, \ldots, a_N)$ with a bilinear softmax model:
\begin{equation}
\label{expr:reconstr}
p(a_i | \bs{a}_{-i},\bs{r}, v, C, \bs{u}) = \frac{\exp({  \bs{u}_{a_i}^T 
C_{v,r_i}^T   \sum_{j \neq i} { C_{v,r_j} \bs{u}_{a_{j}} }})}{ Z(\bs{r}, v, i)},
\end{equation}
 $\bs{u}_{a} \in \mathbb{R}^d$ (for every $a \in \mathcal{A}$)  and $C_{v,r} \in \mathbb{R}^{d \times k}$ (for every verb $v$ and every role $r \in \mathcal{R}$)
 are model parameters, $Z(\bs{r}, v, i)$ is the partition function ensuring that the probabilities sum to one.
Intuitively, embeddings $\bs{u}_a$, when learned from data, will encode semantic properties of an argument: for example, embeddings for the words \mention{demonstrator} and \mention{protestor} should be somewhere near each other in $\mathbb{R}^d$ space, and further away from that for the word \mention{cat}. The product $C_{p,r} \bs{u}_a$ is a $k$-dimensional vector encoding beliefs about other arguments based on the argument-role pair 
$(a, r)$. For example, seeing the argument \mention{demonstrator} in the \entity{Patient} position for the predicate \mention{charge}, one would predict that  the \entity{Agent} is perhaps the word \mention{police}, and the role \entity{Instrument} is filled by the word \mention{baton} or perhaps (a water) \mention{cannon}. On the contrary,
if the \entity{Patient} is  {\it cat} then the \entity{Agent} is more likely to be  {\it dog} than  {\it police}.
In turn, the dot product $(C_{v,r_i} \bs{u}_{a_i})^T C_{v,r_j} \bs{u}_{a_j}$ is large if these expectations are met for the argument pair $(a_i, a_j)$, and small otherwise. 
Intuitively, this objective corresponds to scoring argument tuples according to
\begin{equation}
\vspace{-1ex}
\nonumber
h(\bs{a}, \bs{r}, v, C, \bs{u}) = \sum_{i \neq j} {\bs{u}_{a_i}^T C^T_{v,r_i} C_{v,r_j} \bs{u}_{a_j}},
\end{equation}
hinting at connections to
(coupled) tensor and factorization methods~\cite{nickel2011,yilmaz2011,bordes2011,riedel2013} and distributional semantics~ \cite{mikolov2013efficient,pennington2014}.
Note also that the reconstruction model does not have 
access to any features of the sentence
 (e.g., argument order or syntax), forcing the roles to convey all the necessary information. 
 
 This factorization can be though of as a generalization of the notion of selection preferences. Selectional preferences characterize the set of arguments licensed for a given role of a given predicate: for example, \entity{Agent} for the predicate {\it charge} can be  \mention{police} or  \mention{dog} but not \mention{table} or \mention{idea}. In our generalization, we model soft restrictions imposed not only by the role itself but also by other arguments and their assignment to roles. 
 
 
 
In practice, we extend the model slightly: (1) we introduce a word-specific bias (a scalar $b_a$ for every $a \in \mathcal{A}$) in the argument prediction model 
(equation (\ref{expr:reconstr})); (2) we smooth the model
by using a sum of predicate-specific and cross-predicate projection matrices $(C_{v,r} + C_{r})$ instead of just $C_{v,r}$.


\subsection{Learning}
\label{sect:learning}

Parameters of both model components ($\bs{w}$, $\bs{u}$ and $C$)  are  learned jointly: the 
natural objective associated with every sentence would be the following:
\begin{equation}
\label{expr:unlabobj}
 \sum_{i=1}^N {\log \sum_{\bs{r}} { p(a_i | \bs{a}_{-i}, \bs{r}, v, C, \bs{u}) p(\bs{r} | x, \bs{w})}}.
\end{equation}
However optimizing this objective is not practical in its exact form for two 
reasons: (1) marginalization over $\bs{r}$ is exponential in the number of arguments; (2) the partition
function $Z(\bs{r}, v, i)$ requires summation over the entire set of potential argument lemmas. We use existing techniques
to address both challenges.

In order to deal with the first challenge, 
we use a basic mean-field approximation. Namely, instead of computing an expectation of $p(a_i | \bs{a}_{-i},\bs{r}, \!v, \!C,\! \bs{u})$ under $p(\bs{r} | x, \bs{w})$, as in (\ref{expr:unlabobj}), we use
the posterior distributions  $\mu_{is} = p(\bs{r_i} = s | x, \bs{w})$
and score the argument predictions as 
\begin{align}
\label{expr:softmax}
\!\!& p(a_i | \bs{a}_{-i},\bs{\mu}, \!v, \!C,\! \bs{u}) = \frac{\exp\left(\phi_i(a_i, \bs{a}_{-i}, \bs{ \mu})\right)}{ Z(\bs{\mu}, v, i)} \\ 
\!\!\!& \phi_i(a_i, \bs{a}_{-i},\bs{\mu}) = {\bs{u}_{a_i}^T \! (\sum_{s}
\mu_{is} C_{v,s})^T\!   \sum_{j \neq i} {\!\! (\sum_{s} {\mu_{js}C_{v,s}}) \bs{u}_{a_{j}} }},
\nonumber
\end{align}
where $\bs{\mu}$ are the posteriors for all the arguments, and $ \phi_i(a, \bs{a}_{-i})$ is the score associated with predicting lemma $a$ for the argument $i$. 


In order to address the second problem,  the computation of 
$Z(\bs{\mu}, v, i)$, we use a negative sampling technique (see, e.g.,~\citet{mikolov2013efficient}).
More specfically,  we get rid of the softmax in equation~(\ref{expr:softmax}) and optimize the following  sentence-level objective:
\begin{equation}
\label{expr:objtract}
\! \!\sum_{i=1}^N {\big[ \log   \sigma( \phi_i(a_i,\! \bs{a}_{-i}))\!  -
\! \!\sum_{a' \in S}{ \log  \sigma( \phi_i(a',\! \bs{a}_{-i}))\big]} },
\end{equation}
where $S$ is a random sample of $n$ elements from the unigram distribution of lemmas, and $\sigma$ is the logistic sigmoid function. 

Assuming that the posteriors $\mu$ can be derived in a closed form, the gradients of the objective (\ref{expr:objtract}) with
respect to parameters of both the encoding component ($\bs{w}$)  and the reconstruction component ($C$, $\bs{u}$ and $\bs{b}$) can be computed 
using back propagation. In our experiments, we used the AdaGrad algorithm~\cite{duchi2011adaptive} to perform the optimization.

The learning algorithm is quite efficient, as the reconstruction computation is bilinear, 
whereas the computation of the posteriors $\bs{\mu}$ (and the computation of their gradients) from the semantic roler labeling component (encoder) is not more expensive than discriminative supervised learning of the role labeler.  Moreover,  the computations can be sped up substantially by observing that the
sum $\sum_{s}{\mu_{is} C_{v,s}}$ in the expression (\ref{expr:softmax}) can be
precomputed for all $i$, and reused  across predictions of different arguments of the same predicate.
At test time, only the linear semantic role labeler is used, so the inference is straightforward.

%

\section{Experiments}
\label{sec:exp}

\subsection{Data and evaluation metrics}
\label{ssec:exp-metrics}

We considered English and German in our experiments. 
For each language, we replicated experimental set-ups used in previous work.

For English, we followed~\citet{Lang10} and used the dependency version of
PropBank~\cite{Palmer05} released for the CoNLL 2008 shared task~\cite{conll08}. The dataset is divided
into three segments. As in the previous work on unsupervised role labeling, we
used the largest segment (the original CoNLL training set, sections 2-21) both for evaluation and learning. This is permissible as unsupervised models do not use gold labels in training.
The two small segments (sections 22 and 23) were used for model development.
In our experiments, we relied on gold standard syntax and gold standard argument identification, as this set-up allows us to
evaluate against much of the previous work. 
We refer the reader to~\citet{Lang10} for details of the experimental set-up.

There has not been much work on unsupervised induction of roles for languages other than English, perhaps primarily because 
of the above-mentioned model limitations.
For German, we replicate the set-up considered in~\citet{TitovAcl2012}.
They used the CoNLL 2009 version~\cite{conll09}  of the SALSA corpus~\cite{Burchardt2006}.  
Instead of using syntactic parses provided in the CoNLL dataset, they re-parsed it with the MALT dependency parser~\cite{Nivre04}.
Similarly, rather than relying on  gold standard annotations for argument identification, they used a supervised classifier to predict argument positions.
Details of the preprocessing can be found in~\citet{TitovAcl2012}.  


As in most previous work on unsupervised SRL, we evaluate our model using purity, collocation and their harmonic mean F1. \emph{Purity} (PU)  measures the average number of arguments with the same gold role label in each cluster, \emph{collocation} (CO) measures to what extent a specific gold role is represented by a single cluster. 
More formally:
\begin{equation*}
PU = {1 \over N} \sum_i \max_j |G_j \cap C_i |
\end{equation*}

\noindent where if $C_i$ is the set of arguments in the $i$-th induced cluster,  $G_j$ is the set of arguments in the $j$th gold cluster, and $N$ is the total number of arguments.  Similarly, for collocation: 
\begin{equation*}
CO = {1 \over N} \sum_j \max_i |G_j \cap C_i |
\end{equation*}

We compute the aggregate PU, CO, and F1 scores over all predicates in the same way as~\citet{Lang10} by weighting 
the scores for each predicate by the number of times its arguments occurr.

\subsection{Parameters and features}
\label{ssec:exp-param}

For the semantic role labeling (encoding) component,
we relied on 14 feature patterns used for argument labeling in one of 
popular supervised role labelers~\cite{johansson2008}.
These patterns include non-trivial syntactic features, such as a dependency path between the target predicate and the considered argument.
The resulting feature space is quite large  (49,474 feature instantiations for our English dataset) and
arguably sufficient to accurately capture syntax-semantics interface for most languages.
Importantly, the dimensionality of the feature space is very different from the one used typically in unsupervised SRL. 
In principle, any features could be used here but we chose these 14 feature patterns,
as they all are fairly simple and generic. They can also be easily extracted from any treebank.  We used the same feature patterns both for English and German. However, there is little doubt that some language-specific feature engineering and the use of language-specific priors or constraints (e.g., posterior regularization~\cite{Ganchev10})  would benefit the performance.
Faithful to our goal of constructing the simplest possible feature-rich model,  we use logistic classifiers independently
predicting role distribution for every argument. 

For the reconstruction component, both for English and German, we set the dimensionality of embeddings $d$,   the projection dimensionality $k$ and the number
of negative samples $n$ to $30$, $15$ and $20$, respectively. The model was not sensitive to the parameter defining the number of roles as long it was large enough (see Section~\ref{ssec:exp-res} for more discussion).
For training, we used uniform random initialization and AdaGrad~\cite{duchi2011adaptive}. 
 Any model selections (e.g., choosing the number of epochs) was done on the basis of the respective development set.

\subsection{Results}
\label{ssec:exp-res}

\subsubsection{English}
\label{sssec:exp-res-eng}

 Table~1 
 summarizes the results of our method, as well as 
 those of alternative approaches and baselines.
 
Following ~\cite{Lang10}, we use a baseline ({\em SyntF}) which simply
clusters predicate arguments according to the dependency relation to their head.
A separate cluster is allocated for each of 20 
most frequent relations in the dataset and an additional cluster is used for all
other relations.
As observed in the previous work~\cite{Lang11a}, this is a hard baseline to beat. 

We also compare against previous approaches:
the latent logistic classification model \cite{Lang10}  (labeled {\em LLogistic}), 
the  agglomerative clustering  method \cite{Lang11a} ({\em Agglom}),  the graph partitioning approach \cite{Lang11b} ({\em GraphPart}),
the global role ordering model~\cite{garg2012} ({\em RoleOrdering}). We also report results of an  improved version of {\em Agglom},
recently reported by~\citet{Lang14} ({\em Agglom+}).
The strongest previous model is {\em Bayes}: {\em Bayes} is the most accurate (`coupled') version of  
the Bayesian model of~\citet{TitovKlementEacl12},
estimated from the CoNLL data without relying on any external data.~\citet{TitovKlementEacl12} also showed that using Brown clusters induced  from a large external corpus resulted in an 0.5\% improvement in F1 but that version
 is not entirely comparable to other systems  induced solely from the CoNLL
 text.
 
Our model outperforms or performs on par with best previous models in terms of F1. Interestingly, the purity and collocation balance is very different for our model and for the rest of the systems.
In fact, our model induces at most 4-6 roles.  On the contrary,  {\em Bayes} predicts more than 30 roles for the majority of frequent predicates (e.g., 43 roles for the predicate {\it include} or 35 for {\it say}).
Though this tendency reduces the purity scores for our model, this also means that our roles are more human interpretable. For example, agents and patients are clearly identifiable in the model predictions.  Our model has similar purity to the syntactic baseline but outperforms it vastly according to the collocation metric, suggesting
that we go substantially beyond recovering syntactic relations.

In additional experiments, we observed that our model, in some regimes, starts to induce roles
specific to individual verb senses or  specific to groups of  semantically similar predicates. This suggests that adding a latent variable capturing predicate 
senses and conditioning the reconstruction component on this variable  may not only result in a more informative semantic representation (i.e. include verb senses) but also  improve the role induction performance. We leave this exploration for future work.

\begin{table}
\begin{center}

\begin{tabular} {l c c c}  
\label{tab:en}
                & PU & CO & F1 \\
\hline
Our Model         & 79.7 & 86.2 & {\bf 82.8}   \\
Bayes           & 89.3 & 76.6 & 82.5   \\
Agglom+          &  87.9 & 75.6 & 81.3 \\
RoleOrdering    & 83.5 & 78.5 & 80.9 \\
Agglom        & 88.7 & 73.0 & 80.1   \\
GraphPart		  &	88.6 & 70.7 & 78.6   \\
LLogistic         & 79.5 & 76.5 & 78.0   \\

\hline
SyntF		      & 81.6 & 77.5 & 79.5   \\
\end{tabular}
\caption{Results on English (PropBank / CoNLL 2008).}
\end{center}
\end{table}

\subsubsection{German}
\label{sssec:exp-res-de}

For German, we replicate the experimental set-up previously
used by~\citet{TitovAcl2012}. As for English,
we report results of the syntactic baseline ({\em SyntF}).  
The results for all approaches are presented in Table~2.
We compare against {\em Bayes+LangSpecific} -- the {\em Bayes} model~
with argument signatures specialized for German (as reported in~\citet{TitovAcl2012}).
We also consider the original version of the {\em Bayes} model (denoted as {\em Bayes}).

 Recently,~\citet{Lang14}, evaluated their {\em Agglom+} on a version of the
 same German SALSA dataset.
Their best result is F1 of 79.2\%, however, this score and our results are not directly comparable.  Instead of 
using the CoNLL dataset, they processed the corpus themselves. They also relied on syntactic features from a constituent 
parser whereas dependency representations are used in our experiments.

The overall picture for German closely resembles the one for English. Our method achieves results comparable to
the best method evaluated in this setting. 
Importantly, parameters and features of our model for German and English are identical. 
On the contrary, by comparing  {\em Bayes} with {\em Bayes+LangSpecific},  one can see that specialization of argument
signatures was crucial for the Bayesian model.  Also, similarly to English, our method induces less fine-grain sets of semantic roles but achieves
much higher collocation scores.


\begin{table}
\begin{center}

\begin{tabular}{l*{5}{c}r}
                & PU & CO & F1 \\
\hline
Our Model         & 76.3 & 87.0 & {\bf 81.3}   \\
Bayes+LangSpecific         & 86.8 & 75.7 & 80.9   \\
Bayes & 80.6 & 76.0 & 78.3 \\
\hline
SyntF		      & 83.1 & 79.3 & 81.2  \\
\end{tabular}
\end{center}
\caption{Results on German (SALSA / CoNLL 2009).}
\end{table}

\section{Additional related work}

In recent years, unsupervised approaches to semantic role induction 
have attracted considerable attention.   However,  there exist other ways to address  
insufficient coverage provided by  existing semantically-annotated resources. 

One natural direction is semi-supervised role labeling, where both annotated and unannotated data is used to construct a model.
Previous  semi-supervised approaches to SRL can mostly be regarded as extensions to supervised learning by either incorporating word features induced from unnannoted texts ~\cite{collobert2008,deschacht2009} or creating some form of `surrogate' supervision~\cite{he2006,Fuerstenau09,das2011}.
The benefits from using unlabeled data were  moderate, and more significant for the harder SRL version,  frame-semantic parsing~\cite{das2011}.

Another important direction includes cross-lingual approaches~\cite{Pado09,vanderPlas11} which leverage resources for research-rich languages, as well as parallel data, to transfer the annotation to the resource-poor languages. However, both translation shifts and noise in word alignments harm the performance of  cross-lingual methods. Nevertheless, even joint unsupervised induction across languages appears to be beneficial~\cite{TitovAcl2012}.

Unsupervised learning has also been one of the central paradigms for 
the closely-related area of relation extraction (RE), where 
several techniques have been proposed to cluster semantically
similar verbalizations of
relations~\cite{Lin01,Banko07,yao2011structured}. Similarly to SRL,
unsupervised methods for  RE mostly rely on generative modeling and agglomerative clustering.

From the learning perspective,  methods which use  the reconstruction-error objective to estimate  linear models \cite{ammar2014,daume2009} are certainly related. However, they do not consider learning factorization models, and they also do not deal with semantics.   
Tensor and factorization methods used in the context of  modeling knoweldge bases
(e.g., \cite{bordes2011}) are also close in spirit. However, they do not deal with inducing semantics but rather factorize existing relations (i.e. rely on semantics).

\section{Conclusions and discussion}

This work introduces a method for inducing feature-rich semantic role labelers from unannoated text.
 In our approach, we view 
a semantic role representation as an encoding of a latent relation between a predicate and a tuple of its arguments. We capture this relation with
a probabilistic tensor factorization model.  The factorization model (relying on semantic roles) and a feature-rich model (predicting the roles) are jointly estimated  by optimizing an objective which favours accurate reconstruction of arguments given the latent semantic representation (and other arguments).  
Our estimation method yields a semantic role labeler which achieves state-of-the-art results both on English and German.

Unlike previous work on role induction,  in our approach, virtually any computationally tractable structured model can be used as the role labeler , including almost any
semantic role labeler introduced in the context of  supervised SRL (see, e.g., CoNLL shared tasks ~\cite{conll05,conll08,conll09}). This opens interesting
possibilities  to extend our approach to the semi-supervised setting.  
Previous unsupervised SRL models were making too strong assumption and used too
limited features to effectively use unlabeled data. For our model, the reconstruction objective can be easily combined with the likelihood objective, yielding a potentially powerful semi-supervised method. We leave this  direction for future work.


\section*{Acknowledgments}

The work was partially supported by a Google Focused Award on Natural Language
Understanding. The authors thank Dipanjan Das, Mikhail Kozhevnikov, Ashutosh
Modi and Alexis Palmer for insightful suggestions.

\bibliography{ivan}
\bibliographystyle{apalike}

\end{document}